\pdfoutput=1

\documentclass[11pt]{article}

\usepackage[final]{acl}

\usepackage{times}
\usepackage{latexsym}

\usepackage[T1]{fontenc}

\usepackage[utf8]{inputenc}

\usepackage{microtype}

\usepackage{inconsolata}

\usepackage{graphicx}

\newcommand{\faithcamera}{\textsf{FaithCAMERA}}
\newcommand{\camera}{\textsf{CAMERA}}
\newcommand{\cameracubed}{\textsf{CAMERA$^3$}}
\newcommand{\pt}{\textrm{prec}_{t}}
\newcommand{\ps}{\textrm{prec}_{s}}

\usepackage{booktabs}

\usepackage{multirow}

\usepackage{amssymb}
\usepackage{tikz}
\usepackage[xcolor,framemethod=tikz]{mdframed}
\usepackage{color}
\usepackage{colortbl}
\usepackage{paralist}

\newcommand{\bgfb}[1]{\tikz[baseline=(X.base)]{\node(X)[rectangle, fill=red!20!white, rounded corners, text height=.8ex,text depth=-0.5ex]{\textit{#1}};}}

\DeclareRobustCommand{\testbgfa}[1]{\tikz[baseline=(X.base)]{\node(X)[rectangle, fill=cyan!60!blue!18, rounded corners, text height=.8ex,text depth=-0.5ex]{\textit{#1}};}}
\DeclareRobustCommand{\testbgfb}[1]{\tikz[baseline=(X.base)]{\node(X)[rectangle, fill=red!20!white, rounded corners, text height=.8ex,text depth=-0.5ex]{\textit{#1}};}}
\DeclareRobustCommand{\testbgfc}[1]{\tikz[baseline=(X.base)]{\node(X)[rectangle, fill=green!20!white, rounded corners, text height=.8ex,text depth=-0.5ex]{\textit{#1}};}}

\usepackage[whole]{bxcjkjatype}

\title{\faithcamera: Construction of a Faithful Dataset for Ad Text Generation}

\author{
    Akihiko Kato, Masato Mita, Soichiro Murakami, Ukyo Honda, Sho Hoshino, Peinan Zhang\\
    Cyberagent. Inc.\\
    \texttt{\{kato\_akihiko, mita\_masato, murakami\_soichiro\}@cyberagent.co.jp}\\
    \texttt{\{honda\_ukyo, hoshino\_sho, zhang\_peinan\}@cyberagent.co.jp}
}

\begin{document}
\maketitle
\begin{abstract}
In \textit{ad text generation (ATG)}, desirable ad text is both faithful and informative. That is, it should be faithful to the input document, while at the same time containing important information that appeals to potential customers.
The existing evaluation data, \camera~\citep{mita-etal-2024-striking}, is suitable for evaluating informativeness, as it consists of reference ad texts created by ad creators.
However, these references often include information \textit{unfaithful} to the input, which is a notable obstacle in promoting ATG research.
In this study, we collaborate with in-house ad creators to refine the \camera~references and develop an alternative ATG evaluation dataset called \faithcamera, in which the faithfulness of references is guaranteed.
Using \faithcamera, we can evaluate how well existing methods for improving faithfulness can generate informative ad text while maintaining faithfulness.
Our experiments show that removing training data that contains unfaithful entities improves the faithfulness and informativeness at the entity level, but decreases both at the sentence level.
This result suggests that for future ATG research, it is essential not only to scale the training data but also to ensure their faithfulness.
Our dataset will be publicly available~\footnote{\url{https://github.com/CyberAgentAILab/FaithCAMERA} \label{fot:repo}}.
\end{abstract}

\section{Introduction}
\label{sec:intro}
\begin{figure}[t]
 \centering
  \includegraphics[width=0.9\linewidth]{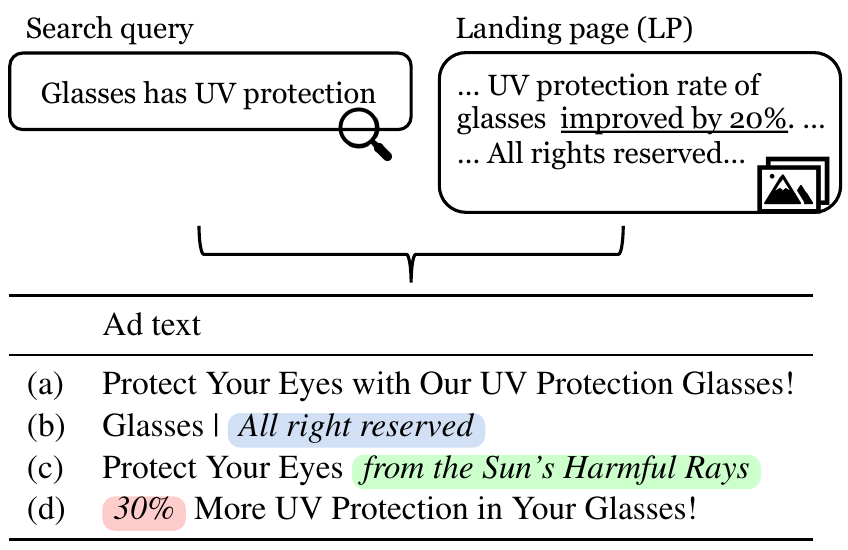}
  \caption{Overview of ad text generation.\\
  The highlighted areas in each color indicate 
  \testbgfa{faithful and uninformative}, \testbgfc{factual but unfaithful} \\and \testbgfb{non-factual and unfaithful} spans, respectively. Note that the quantity listed in (d) (``30\%'') is different from what is written on the LP (``20\%'').}
 \label{fig:example_motivation}
\end{figure}

The increasing demand for large-volume ad production driven by the spread of online advertising has led to a growing interest in research focused on the automatic generation of ad texts, often referred to as \textit{ad text generation}~\citep[ATG;][]{Bartz-et-al:08,Fujita:10,Hughes-et-al:19,Youngmann-etal:20,kamigaito-etal-2021-empirical,golobokov-etal-2022-deepgen}.
Here, ATG is defined as a task that outputs appealing ad texts given source documents such as landing pages (LPs) and user signals such as search queries as inputs (Figure~\ref{fig:example_motivation}).
\begin{figure*}[t]
 \centering
  \includegraphics[width=0.9\linewidth]{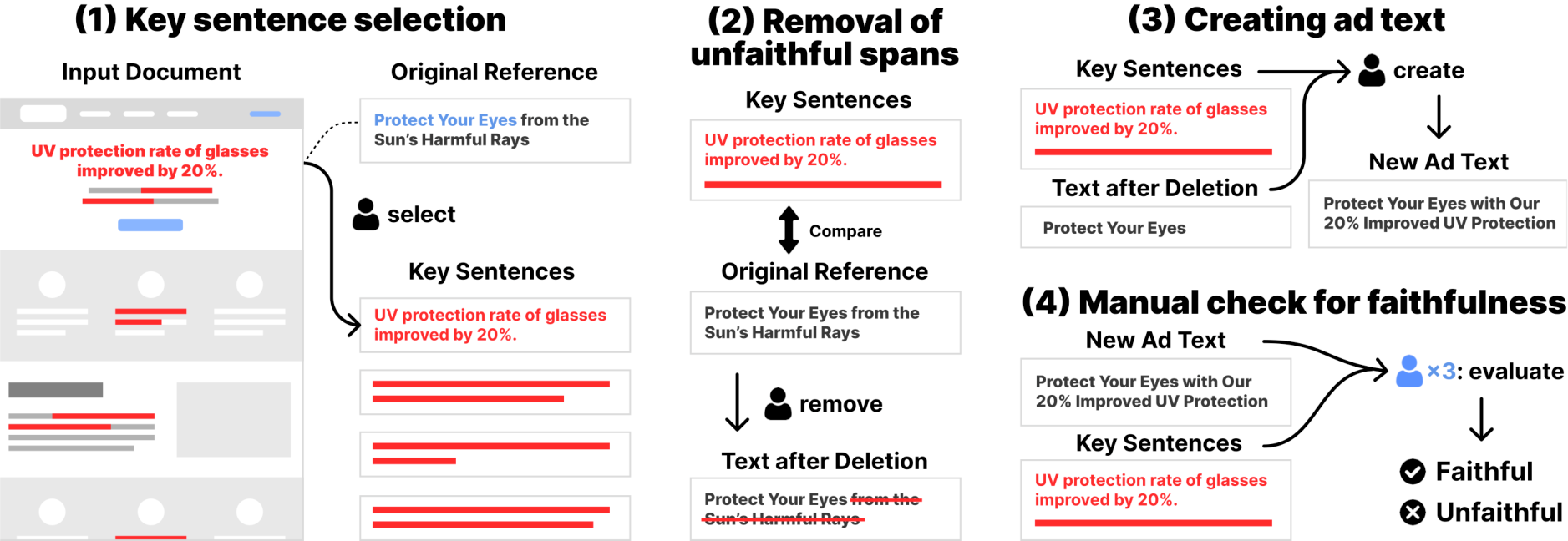}
 \caption{Overview of dataset construction flow.}
 \label{fig:flow}
\end{figure*}

According to \citet{mita-etal-2024-striking}, who first explicitly defined ATG as an NLP task, ad text should be faithful to the input documents (\textit{faithfulness}\footnote{Although definitions vary across studies, this study follows~\citet{maynez-etal-2020-faithfulness}: an ad text is considered \textit{unfaithful} if it contains any span not semantically supported by the input document. See Appendix \S\ref{sec:appendix-h} for a discussion of why faithfulness is important in the ATG and the relationship between faithfulness and factuality.}) and include key information appealing to potential customers (\textit{informativeness}).
However, developing ATG models that ensure both faithfulness and informativeness has been challenging in the current situation.
While \camera~\cite{mita-etal-2024-striking}, the first publicly available benchmark dataset in the ATG field, and its variant \cameracubed~\cite{inoue-etal-2024-camera3-evaluation}, which ensures greater diversity of appeals, are suitable for evaluating informativeness as they consist of reference ad texts created by ad creators, these references often contain \textit{unfaithful} information, such as entities that do not appear in the input\footnote{\citet{mita-etal-2024-striking}~demonstrates this quantitatively using \textit{novel} n-grams in their own paper.}(Figure~\ref{fig:example_motivation}(c)).
If the faithfulness of references is not guaranteed, it is not possible to evaluate whether the generated ad text is excellent in terms of both informativeness and faithfulness using a single reference-based metric. Therefore, the lack of evaluation data with guaranteed faithfulness is a notable obstacle to promoting ATG research.

To address this obstacle, we collaborate with in-house ad creators to construct \faithcamera, a Japanese ATG evaluation dataset that focuses on faithfulness while ensuring informativeness, by refining the~\camera~references.

Using~\faithcamera, we evaluate the ability of existing faithfulness-improvement methods to produce informative ad text while maintaining faithfulness, and find that \textit{data filtering}~\cite{nan-2021} increases faithfulness and informativeness at the entity level, while decreasing both at the sentence level. This result suggests that, in future ATG research, it is essential to not only increase the size of the training data, but also to ensure its faithfulness.
Our dataset will be publicly available~\footref{fot:repo}.

\section{Construction of \faithcamera}
\label{sec:corpus-construction}
This study aims to construct an evaluation dataset with gold references that are faithful to the input by refining the references of an existing evaluation dataset, \camera.
As refining strategies, there are two possible approaches to creating faithful ad texts:
\begin{inparaenum}[(a)]
\item binary classification of \camera~references as faithful or not and extraction of a faithful subset (\textit{filtering approach}), or
\item editing \camera~references (\textit{editing approach}).
\end{inparaenum}
In this study, we adopt the editing approach because the filtering approach would not ensure a sufficient amount of test data (\S\ref{subsec:procedure}), and there is a concern that the distribution of the data may be biased.
In fact, the editing approach has been employed in related works of other natural language generation (NLG) tasks, including data-to-text~\citep{parikh-2020} and dialogue~\citep{dziri-2022}.

\subsection{Data Source}
We use the test split of \camera~\cite{mita-etal-2024-striking} as source data.
Each test instance consists of an input \textit{x} and reference ad texts \textit{y}. 
\textit{x} consists of (1) a search query, (2) a description of the LP, and (3) the text of the OCR-processed LP full view. 
\textit{y} consists of a delivered reference and three additional references, which are created by three expert annotators.
To construct \faithcamera, we treat sentences obtained by concatenating the above (1)-(3) as input information. 
In addition, we use the above delivered reference as a reference ad text for refinement. 
The reason for using only the delivered reference is that the purpose of \faithcamera~construction is not to create a ``dataset that can replace \camera,'' but to provide the research community with an evaluation dataset that complements \camera.

\subsection{Construction Procedure}
\label{subsec:procedure}
An overview of our four-step dataset construction procedure is shown in Figure~\ref{fig:flow}. 
Annotation is performed by annotators with experience in in-house ad production.
One annotator is assigned to each case for steps (1)-(3) below, and three annotators are assigned to each case for step (4), which involves manually checking the faithfulness of the reference text.
Each step of the annotation process is described below:

\paragraph{(1) Key sentence selection:}
Select key sentences (the red text in Figure~\ref{fig:flow}) that support an original reference from an input document and create a set of key sentences.

\paragraph{(2) Removal of unfaithful spans:}
Remove spans containing information not in the set of key sentences (i.e., spans of \textit{factual hallucination}~\cite{cao-etal-2022-hallucinated}, henceforth \textit{unfaithful spans}) from the original reference.
In the cases where no unfaithful spans are present (199 out of 872 cases), we skip the following steps (3) and (4) and adopt the original reference as the faithful gold reference.

\paragraph{(3) Creating ad text:}
Based on the modified reference, in which unfaithful spans are removed, an annotator creates a new reference to be faithful to the input, referring to the set of key sentences.
In cases where no key sentences are selected in step (1), the annotator is instructed to consider the entire input information as a set of key sentences (13 out of 872 cases).

\paragraph{(4) Manual check for faithfulness:}
For the new reference edited to be faithful to the input in step (3), three annotators evaluate whether the new reference is faithful, referring to the set of key sentences.
For each test instance, these three annotators are different from the annotator who edits the ad text in step (3).
Cases evaluated as \textit{unfaithful} by a majority vote are returned to step (3) and re-edited.
These steps are repeated until all cases are evaluated as \textit{faithful}.

Considering that reference sentences in~\camera~are not faithful to the input in 673 of 872 cases (77.1\%) ,~\faithcamera~shows a significant improvement in faithfulness compared to~\camera. Besides, based on the assumption that human-generated references contain important information that appeals to potential customers, we can expect that the key sentences selected in step (1) contain important information. Therefore, step (3) yields a new reference that is faithful and informative.

\paragraph{To what extent do annotations for faithfulness judgments vary among annotators?}

In step(4), of the 673 cases in which the~\camera~reference ad text is modified in step (2) to ensure faithfulness, all annotators in 671 cases (99.7\%) and two out of three annotators in the remaining two cases judge the modified reference ad text to be faithful to the input.

\subsection{Statistical Analysis}
\label{subsec:stat-analysis}
We investigate whether the construction procedure described in~\S\ref{subsec:procedure} results in~\faithcamera~being overly extractive and whether the task of generating faithful and informative ad text is too easy in~\faithcamera. As a result, we find that~\faithcamera~is more extractive than~\camera, but not overly extractive. See Appendix \S\ref{sec:appendix-e} for details. This finding suggests that simple extractive summarization is insufficient to generate faithful and informative ad text in~\faithcamera.

\section{Model Evaluation on  \faithcamera}
\label{sec:experiments}
To demonstrate the usefulness of~\faithcamera, we investigate how effective the two major existing faithfulness improvement methods (\S\ref{subsec:faithful_methods}) are in ATG when evaluated with~\faithcamera.

We use~\textrm{mT5}\_\textrm{base}\footnote{\url{https://huggingface.co/google/mt5-base}}~\cite{xue-etal-2021-mt5} as a base pre-trained model and fine-tune it on the training and development split of CAMERA\citep{mita-etal-2024-striking} to create the base ATG model for applying the faithfulness improvement methods. Other details are described in Appendix \S\ref{sec:appendix-a}.

We mainly focus on \textit{entity-level} faithfulness because commercial risks are significant if the faithfulness of entities is not ensured. 
In fact, named entities such as product names (e.g., \textit{iPhone 15 Plus}) and numerical expressions such as discounted prices (e.g., \textit{50\% off}) are used in advertisements to make effective appeals~\cite{murakami-etal-2022-aspect}.

\begin{table*}[t]
    \centering
    \scalebox{0.75}{
    \begin{tabular}{l|ccc|ccccc}
        \toprule
        \multirow{3}{*}{\textbf{Models}}&\multicolumn{3}{c|}{\textbf{Automatic Evaluation}}&\multicolumn{5}{c}{\textbf{Manual Evaluation}}\\
        & $\ps$ & $\pt$ & ROUGE-L &\underline{Faithful ($\uparrow$)}&\multicolumn{4}{c}{\underline{Unfaithful ($\downarrow$)}}\\
       &  & & &  &Ent.& Symb. & Pred. & Oth.   \\ \hline
        Baseline &68.6 $\pm$ 0.4 & 21.1 $\pm$ 0.4 & \textbf{29.1 $\pm$ 0.3} & 46 / 100 & 25 & 22  & 15 & 6\\
        Data filtering & \textbf{79.0} $\pm$ 1.2 & \textbf{23.5 $\pm$ 0.4} & 27.6 $\pm$ 0.4 & 59 / 100 & \textbf{15} & \textbf{14} & 16 & 10 \\
        Loss truncation & 71.0 $\pm$ 1.4 & 20.1 $\pm$ 1.0 & 28.1 $\pm$ 0.9 & \textbf{62} / 100  & 26 & 23 & \textbf{4} & \textbf{2} \\
        \bottomrule
    \end{tabular}
    }
    \caption{Automatic and manual evaluation results: a \textbf{bold} value indicates the best result and each value in automatic evaluation for models is the average and standard deviation of three runs with different seeds. The total number of unfaithful cases across models does not match since each model output may contain more than one type of unfaithful spans.}
    \label{tab:main_result}
\end{table*}

\subsection{Methods for Improving Faithfulness}
\label{subsec:faithful_methods}
\paragraph{Data filtering~\cite{nan-2021}}
A method of model training after filtering out training cases whose output contains entities~\footnote{We have targeted named entities, time expressions, terms, and numerical expressions for extraction.} that are not faithful to the input. Details are described in Appendix \S\ref{sec:appendix-f}.

\paragraph{Loss truncation~\cite{kang-2020}}
A method that dynamically truncates high-loss cases~\footnote{Following~\citet{kang-2020}, we set the hyper parameter~\texttt{dropc} to 0.4.~\texttt{dropc} specifies how many percentages of the top percentile of case losses in a mini-batch should be truncated.} in a mini-batch in the gradient calculation during training.

\subsection{Evaluation}
\label{subsec:evaluation}
\paragraph{Automatic evaluation}
To evaluate entity-level faithfulness and informativeness, we use the precision-target($\pt$)~\cite{nan-2021}, which is a reference-based evaluation metric calculated as the percentage of entities in a model output that also appear in the gold reference. 
For evaluations that focus solely on faithfulness, we use the precision-source($\ps$)~\cite{nan-2021}, which is a reference-free evaluation metric calculated as the percentage of entities in a model output that also appear in the input document. 
The target entities and their extraction methods follow the settings of \citet{mita-etal-2024-striking}.
Besides, we use ROUGE-L~\citep{lin-2004-rouge}, an evaluation measure based on the n-gram overlap between generated and reference ad texts. Because~\faithcamera~guarantees the faithfulness and informativeness of the reference text, ad texts with high ROUGE-L scores are excellent in terms of both faithfulness and informativeness. Using ROUGE-L not only allows us to evaluate faithfulness and informativeness at the entity level, but also enables us to take into account modifiers of entities and relations between entities in evaluation.

\paragraph{Manual evaluation}
In addition to the automatic evaluation, we conduct a manual evaluation. Specifically, 300 model outputs (3 models $\times$ 100 instances) are randomly selected, and the faithfulness of each model output to the input is determined by majority vote based on the annotation results of three annotators. For error analysis of \textit{unfaithful} cases, we further manually classify the types of unfaithfulness into four categories: entity, symbol, predicate, and other.

\subsection{Results and Discussion}
\label{subsec:discussion}
First, data filtering yields the best $\pt$, 2.4 points higher than the baseline (Table~\ref{tab:main_result}), while ROUGE-L is 1.5 points lower than the baseline.
These results suggest that data filtering improves informativeness and faithfulness at the entity-level, however, it decreases both at the sentence-level.
This is thought to be due to the fact that data filtering reduced the amount of training data by about half (from 12,395 to 5,878 cases). 
Data filtering also yields $\ps$ more than 10 points higher than the baseline. This result suggests that the model is less likely to produce unfaithful entities during inference due to the removal of instances with unfaithful entities from the training data. In fact, in the manual evaluation, data filtering produces fewer \bgfb{unfaithful entities} (e.g., \textit{``If you are looking for administrative jobs with \bgfb{no weekends off} and no outside work''}) than the other methods.

Second, Loss truncation is 1.0 point below baseline for both $\pt$ and ROUGE-L. These results suggest that not only outliers but also instances that are useful for training could be truncated from mini-batches because of their high loss compared to other training instances. On the other hand, loss truncation improves $\ps$ by 2.4 points relative to the baseline, although not as much as data filtering. See Appendix \S\ref{sec:appendix-g} for detailed discussion on loss truncation.

Third, we confirm that the system rankings for entity-level faithfulness are consistent between manual and automatic evaluations. For more details, see Appendix \S\ref{sec:appendix-i}.

Furthermore, we also verify the effectiveness of the extractive summarization method, specifically BM25 in~\faithcamera~(see Appendix \S\ref{sec:appendix-j} for details).
As a result, although BM25 significantly outperforms other methods in $\ps$ and manual evaluation of faithfulness, it is 11.6 and 17.7 points below baseline for $\pt$ and ROUGE-L, respectively. These results support the hypothesis stated in \S\ref{subsec:stat-analysis} that applying a simple extractive summarization approach to~\faithcamera~would be insufficient to generate ad texts that are faithful and informative.

\section{Conclusion}
We worked on ATG that achieves both faithfulness and informativeness.
To properly assess the degree to which this goal has been achieved, we constructed \faithcamera, the first Japanese evaluation dataset with guaranteed faithfulness.
Through experiments, we found that data filtering~\cite{nan-2021} improves faithfulness and informativeness at the entity level, while decreasing both at the sentence level. This result suggests that for future ATG research, it is essential not only to scale the training data but also to ensure their faithfulness.


\section*{Limitations}
In this study, we created an evaluation dataset (\faithcamera) that guaranteed the faithfulness of reference ad texts, and evaluated the ability of existing faithfulness improvement methods to create informative ad text while maintaining faithfulness. As a result, we found that data filtering focusing on unfaithful entities increased faithfulness and informativeness at the entity level, but decreased both at the sentence level. Considering that data filtering reduced the training data by about half, future research could focus on augmenting the training data by synthesizing high-quality references that do not contain unfaithful spans with large language models.

\section*{Ethics Statement}
One ethical concern is that models trained on data that contains unfaithful references tend to generate unfaithful ad texts, which means there is a risk of providing users with non-factual information. This could have serious consequences for users and advertisers. Therefore, it is important to develop technology to generate ad text that is faithful to the input, such as landing pages. 

\bibliography{main}

\appendix

\section{Faithfulness in ATG, and its relationship to factuality}
\label{sec:appendix-h}
The reasons why faithfulness is important in ATG are discussed below.
It is essential for advertising effectiveness that an ad text be informative. On the other hand, if an ad text is factual, faithfulness need not necessarily be ensured.
However, a model trained on an ATG dataset without any special consideration for faithfulness is likely to produce unfaithful output and may generate an ad text that is not based on facts (Figure~\ref{fig:example_motivation}-c). This is because ad datasets often contain reference ad texts that are factual but unfaithful to the input (Figure~\ref{fig:example_motivation}-b). Ad texts written by ad creators often incorporate information that is not in the input, specifically, common sense or external knowledge, to enhance the promotion of a product or service~\citep{mita-etal-2024-striking}.

Based on the above, there are two possible strategies to enhance the factuality of ATGs: (1) Learning a factual, although not faithful ATG model, or (2) Learning a faithful ATG model.
To achieve (1), in each training instance, if the reference ad text contains information that is not in the input, it is necessary to find the information in question from an external source that is different from the ad dataset. However, when the ad dataset contains examples about products from various industries, the search for unfaithful information is itself a challenging task, since the information sources to be searched for differ by industry. With the above background, this study regards the realization of (1) as a future work and addresses (2) as an important step to achieve factuality.

\section{Extractiveness of~\faithcamera}
\label{sec:appendix-e}
\begin{table}[t]
\centering
\small
\begin{tabular}{lrr}
\toprule
& \faithcamera & CAMERA\\ \midrule
Coverage & 0.97 $\pm$ 0.10  & 0.81  $\pm$ 0.22\\ 
Density  & 3.16 $\pm$ 2.09 & 2.07 $\pm$  2.15\\
\bottomrule
\end{tabular}
\caption{Coverage and density distributions.}
\label{tab:corpus-stat}
\end{table}
\begin{table*}[t]
    \centering
    \scalebox{0.75}{
    \begin{tabular}{l|ccc|ccccc}
        \toprule
        \multirow{3}{*}{\textbf{Models}}&\multicolumn{3}{c|}{\textbf{Automatic Evaluation}}&\multicolumn{5}{c}{\textbf{Manual Evaluation}}\\
        & $\ps$ & $\pt$ & ROUGE-L &\underline{Faithful ($\uparrow$)}&\multicolumn{4}{c}{\underline{Unfaithful ($\downarrow$)}}\\
       &  & & &  &Ent.& Symb. & Pred. & Oth.   \\ \hline
        Baseline &68.6 $\pm$ 0.4 & 21.1 $\pm$ 0.4 & \textbf{29.1 $\pm$ 0.3} & 46 / 100 & 25 & 22  & 15 & 6\\
        Data filtering & 79.0 $\pm$ 1.2 & \textbf{23.5 $\pm$ 0.4} & 27.6 $\pm$ 0.4 & 59 / 100 & 15 & 14 & 16 & 10 \\
        Loss truncation & 71.0 $\pm$ 1.4 & 20.1 $\pm$ 1.0 & 28.1 $\pm$ 0.9 & 62 / 100  & 26 & 23 & 4 & 2 \\
BM25 & \textbf{96.3} & 9.5 & 11.4 & \textbf{100} / 100 & \textbf{0} & \textbf{0} & \textbf{0} & \textbf{0}\\
        \bottomrule
    \end{tabular}
    }
    \caption{Automatic and manual evaluation results: a \textbf{bold} value indicates the best result and each value in automatic evaluation for models other than extractive summarization (BM25) is the average and standard deviation of three runs with different seeds. Since BM25 is not a neural network-based method and does not have randomness, we describe the results of a single run. The total number of unfaithful cases across models does not match since each model output may contain more than one type of unfaithful spans.} 
    \label{tab:bm25}
\end{table*}
Although the creation procedure in \S\ref{subsec:procedure} ensured that \faithcamera's ad texts are faithful and informative, there is a concern that this may result in an overly extractive problem setting.
In other words, the following question arises: \textit{are ad texts in \faithcamera~completely extractive, or are they somewhat abstract by rearranging and paraphrasing phrases in the input documents?} If the former is true, a simple extractive summarization approach would generate ad text that is faithful to the input and similar to the reference.

To answer the question, we quantify how extractive the ad texts in \faithcamera~are, using two measures of extractiveness commonly used in document summarization: \textit{coverage} and \textit{density}~\citep{grusky-2018}.
Following~\citet{grusky-2018}, we refer to a shared sequence of tokens in an input document and an ad text as an extractive fragment. The coverage measures the extent to which words in the ad text are covered by the set of extractive fragments, and the density measures the average length of the extractive fragments. 

As a result, we find that~\faithcamera~is higher on both measures of extractiveness than~\camera.
On the other hand, given that the mean density is 3.16 for \faithcamera's average ad text length of 14.7 tokens,~\faithcamera~is not overly extractive.
Other evidence in support of this claim includes (i) 38.4\% of the bigrams in ad texts of \faithcamera~do not appear on the input side, and (ii)~\faithcamera~contains many paraphrases such as \textit{``12 hours'' }$\rightarrow$ \textit{``half a day''}.

We show the coverage and density distributions in the both ~\faithcamera~and~\camera in Table~\ref{tab:corpus-stat}. As you can see,~\faithcamera~is higher on both measures of extractiveness than~\camera.
On the other hand, given that the mean density is 3.16 for \faithcamera's average ad text length of 14.7 tokens,~\faithcamera~is not overly extractive.

\section{Details of entity extraction in data filtering}
\label{sec:appendix-f}
We automatically extract entities including named entities (by \texttt{GiNZA}\footnote{\url{https://megagonlabs.github.io/ginza/}}), time expressions (by \texttt{ja\_timex}\footnote{\url{https://github.com/yagays/ja-timex}} ), terms (by \texttt{pytermextract}\footnote{\url{http://gensen.dl.itc.u-tokyo. ac.jp/pytermextract}}), and numerical expressions (by \texttt{pynormalizenumexp}\footnote{\url{https:// pypi.org/project/pynormalizenumexp}}) from the training data, and exclude cases in which these entities appear only in reference. This data filtering operation results in roughly cutting the amount of training data by half (from 12,395 to 5,878).

\section{Implementation details}
\label{sec:appendix-a}
We set the the number of training epochs to 30 and used early stopping by loss in the development set.
The batch size was set to 8, and the maximum input and output lengths were 512 and 128 tokens, respectively.
The GPU used for training was a single A100 40GB.
For implementing the loss truncation, we used the authors' official implementation\footnote{\url{https://github.com/ddkang/loss_dropper}}.

\section{Discussion of evaluation results for loss truncation with respect to precision-source}
\label{sec:appendix-g}
We show experimental results in Table~\ref{tab:main_result}.
Loss truncation improves $\ps$ by 2.4 points compared to baseline, although not as much as data filtering. 
In the manual evaluation, loss truncation produces fewer \bgfb{unfaithful predicates} (e.g., \textit{``Video editing you can \bgfb{learn} while you \bgfb{work}''}) than the other methods. loss truncation is a method to exclude high-loss cases in a mini-batch from the gradient calculation, which reduces the negative impact of noisy training cases (\textit{invalid references}~\cite{kang-2020}) on learning. The above experimental results are consistent with this expectation.

\section{Performance of the extractive summarization model in~\faithcamera}
\label{sec:appendix-j}
We examine the effectiveness of extractive summarization methods in~\faithcamera~to demonstrate the claim stated in~\S\ref{subsec:stat-analysis}. We use the BM25 algorithm~\cite{robertson-2009}. Specifically, keywords are used as queries, and the ad text is created by selecting sentences from (1) the LP description and (2) the OCR-processed LP text in descending order of BM25 score within a range not exceeding the maximum length~\footnote{In accordance with the guidelines for Headline Text in Google Responsive Search Ads (https://support.google.com/ google-ads/answer/12437745), we set this maximum length at 15 full-width characters.} of the ad text.

The results of the BM25 experiment are shown in Table~\ref{tab:bm25}, together with the results of the faithfulness improvement methods described in~\S\ref{subsec:faithful_methods}.

As mentioned in \S\ref{subsec:discussion}, BM25 achieves higher faithfulness than other methods, but it is lower than other methods in terms of informativeness. These results support the hypothesis mentioned in \S\ref{subsec:stat-analysis}, namely that applying a simple extraction summary approach to \faithcamera ~is not sufficient to generate faithful and informative ad texts.

\section{Comparison between the results of automatic and manual evaluation of entity-level faithfulness}
\label{sec:appendix-i}
Regarding human evaluation, we can get the system ranking of entity-level faithfulness by focusing on unfaithful entities. The ranking is as follows: Data filtering > Baseline > Loss truncation. For automatic evaluation, we can evaluate entity-level faithfulness with $\ps$. Here, we focus on 100 instances subjected to manual evaluation. Then, $\ps$ for these instances is calculated as follows: Baseline: 71.1, Data filtering: 76.5, Loss truncation: 68.0. Therefore, the system ranking of faithfulness is Data filtering > Baseline > Loss truncation. By comparing the above two rankings, we can see that the system ranking of entity-level faithfulness is consistent between the automatic and manual evaluation~\footnote{The system ranking of $\ps$ for the entire test data is Data filtering > Loss truncation > Baseline, as shown in Table~\ref{tab:main_result}.}.

\end{document}